# GRAY LEVEL IMAGE ENHANCEMENT USING POLYGONAL FUNCTIONS

**Vasile Pătraşcu**

Department of Informatics Technology
TAROM Company
e-mail: vpatrascu@tarom.ro

**Abstract.** This paper presents a method for enhancing the gray level images. This method takes part from the category of point transforms and it is based on interpolation functions. The latter have a graphic represented by polygonal lines. The interpolation nodes of these functions are calculated taking into account the statistics of gray levels belonging to the image.

## 1. INTRODUCTION

Image enhancement is an important task in the field of image processing. Lengthways time, there were built many methods in this purpose [1], [2], [9], [10]. The great number of existing methods is determined by the great variety of images, which need specific methods [4]. This paper describes a method of image enhancement, which takes part in the group of point transforms [3]. The functions used for gray level transformation are linear interpolation functions on compact intervals from the definition domain. These functions have a graphic represented by polygonal lines while the interpolation nodes are determined out of statistical reasons. The following part of the article is organized thus: section 2 comprises the mathematical theory presentation for the linear interpolation functions on intervals (linear on certain parts of the definition domain); section 3 contains the computing algorithm of the interpolation nodes. The experimental results are presented in section 4, while section 5 offers several conclusions, and finally, the references.

## 2. THE DETERMINATION OF INTERPOLATION FUNCTIONS FOR POLYGONAL TYPE

Let consider the function:

$$f : [0,M] \to [0,M], \qquad f(v) = \sum_{i=1}^{n} a_i |v - v_i| \qquad (1)$$

where $M \in (0,\infty)$. This function depends on the real parameters $(a_i)_{i=1,n}$ and $(v_i)_{i=1,n}$. Let be $v_1 < v_2 < ... < v_{n-1} < v_n$ and $f_1, f_2, ..., f_{n-1}, f_n$ two sets of values from $[0,M]$. The elements $(f_i)_{i=1,n}$ are the function values in the interpolation points $(v_i)_{i=1,n}$. One determines the coefficients $a_1, a_1, ..., a_n$ from the conditions:

_______________________________________________________________________

$$f(v_i) = f_i \qquad \text{with } i = 1,2,...,n \tag{2}$$

The value of the function $f$ in the point $v_i$ implies the expression:

$$f(v_i) = \sum_{j=1}^{i-1} a_j (v_i - v_j) + \sum_{j=i+1}^{n} a_j (v_j - v_i) \tag{3}$$

From relation (3) one infers the following formulae:

$$f(v_i) - f(v_{i-1}) = \left( \sum_{j=1}^{i-1} a_j - \sum_{j=i}^{n} a_j \right)(v_i - v_{i-1}) \quad \text{for } i = 2,...,n \tag{4}$$

and

$$f(v_n) + f(v_1) = \left( \sum_{j=1}^{n} a_j \right)(v_n - v_1) \tag{5}$$

Using (4) and (5) the system of equations (2) becomes:

$$\begin{cases} \sum_{j=1}^{i-1} a_j - \sum_{j=i}^{n} a_j = \dfrac{f_i - f_{i-1}}{v_i - v_{i-1}} & for \quad i = 2,...n \\[2mm] \sum_{j=1}^{n} a_j = \dfrac{f_n + f_1}{v_n - v_1} \end{cases} \tag{6}$$

The system (6) has the subsequent solution:

$$\begin{cases} a_1 = \dfrac{1}{2}\left( \dfrac{f_n + f_1}{v_n - v_1} + \dfrac{f_2 - f_1}{v_2 - v_1} \right) \\[3mm] a_i = \dfrac{1}{2}\left( \dfrac{f_{i+1} - f_i}{v_{i+1} - v_i} - \dfrac{f_i - f_{i-1}}{v_i - v_{i-1}} \right) & for \quad i = 2,...,n-1 \\[3mm] a_n = \dfrac{1}{2}\left( \dfrac{f_n + f_1}{v_n - v_1} - \dfrac{f_n - f_{n-1}}{v_n - v_{n-1}} \right) \end{cases} \tag{7}$$

The relations (7) supply the coefficients values $(a_i)_{i=1,n}$ in determining the function $f$ when knowing the values $(f_i)_{i=1,n}$ in the points $(v_i)_{i=1,n}$.

The graphic of function $f$ is a polygonal line (see Fig.1d, 1f and Fig.2d, 2f).

## 3. THE FIXING OF THE INTERPOLATION NODES

### 3.1 Problem formulation

A gray level image is defined on a compact spatial domain $D \subset R^2$ by a gray level function $l : D \rightarrow [0,M]$ where $[0,M]$ is the gray level set. Usually $M = 255$ or $M = 1$. Assume that the function $l$ is a continuous one and let be $v_1, v_2,...,v_{n-1}, v_n$ the $n$ interpolation points, which must be determined. At the beginning the points $v_1$ and $v_n$ are steady in the values:

$$v_1 = \min_{(x,y) \in D} l(x, y) \tag{8}$$

$$v_n = \max_{(x,y) \in D} l(x, y) \tag{9}$$

The others $n - 2$ points $v_2, v_3,...,v_{n-1}$ will be determined in the next system consisting in

$n - 2$ equations: $\qquad v_i = \dfrac{\displaystyle\int_{D_i} l(x, y) dx dy}{area(D_i)} \qquad \text{for } i = 2,...,n-1 \tag{10}$

where $\quad D_i = \left\{ (x,y) \in D \, / \, l(x,y) \in [v_{i-1}, v_{i+1}] \right\}$ with $i = 2,...,n-1$ (11)

The system (10) yields the solution for the values of interpolation points in the continuous case, and it is useful only from a theoretical point of view.

The practical issues need to consider the discrete case, therefore, the integral operator $\int$ is substituted by the sum operator $\sum$, and $area(D_i)$ by the cardinal $card(D_i)$. Thus, the discrete form of the system (10) is done by:

$$v_i = \frac{\sum\limits_{(x,y) \in D_i} l(x,y)}{card(D_i)} \quad \text{with } i = 2,...,n-1 \quad (12)$$

where the sets $D_i$ are defined in the same way as in the continuous case by (11).

### 3.2 Computation Algorithm

As it is difficult to find an analytical solution for (12), the problem is numerically solved. Suppose that the images' gray levels have discrete values in the interval $[l_{MIN}, l_{MAX}]$. The next procedure will be used for interpolation points calculus:

1. Initialization: choose $n$, $\varepsilon$ the constant for stopping the procedure, $m = 0$, $v_1 = l_{MIN}$, $v_n = l_{MAX}$, and also the initial values $v_2^{(0)} < v_3^{(0)} < ... < v_{n-1}^{(0)}$ so that the sets $D_i^{(0)} = \left\{ (x,y) \in D \, / \, l(x,y) \in [v_{i-1}^{(0)}, v_{i+1}^{(0)}] \right\}$ would verify $D_i^{(0)} \neq \Phi$.

2. For $i = 2,...,n-1$ one calculate $D_i^{(m)} = \left\{ (x,y) \in D \, / \, l(x,y) \in [v_{i-1}^{(m)}, v_{i+1}^{(m)}] \right\}$ and

$$v_i^{(m+1)} = \frac{\sum\limits_{(x,y) \in D_i^{(m)}} l(x,y)}{card(D_i^{(m)})} \quad (13)$$

3. If $\left| v_i^{(m+1)} - v_i^{(m)} \right| < \varepsilon$ pass to the step 4, else $m = m+1$ and go to step 2.

4. Save the results and stop.

Let remark that the described algorithm has the following property: it gives for the interpolation function, a set of equidistant points in the case of a gray level image with an uniform distribution.

## 4. EXPERIMENTAL RESULTS

In order to get some practical results, the proposed method was used to enhance some dark or bright images. The interpolation nodes are chosen so that the new image has a gray level distribution close to an uniform one. Thus, if $[0,M]$ is the interval of gray levels then the values $f_i$ of the interpolation function are equidistant and yield from:

$$f_i = \frac{i-1}{n-1} \cdot M \qquad \text{for } i = 1,2,...,n \quad (14)$$

To exemplify, two images were picked out: one dark ("landsat") in Fig.1a and one bright ("cells") in Fig.2a. Their histograms are in Fig.1b and Fig.2b.

The image "landsat" has the following interpolation functions for $n = 3$ and $n = 4$:

$$f_3(v) = 2.711 \cdot |v - 15| - 0.844 \cdot |v - 53.9| + 0.276 \cdot |v - 134|$$

$$f_4(v) = 2.385 \cdot |v - 15| + 1.712 \cdot |v - 47.4| - 2.44 \cdot |v - 61.4| + 0.486 \cdot |v - 134|$$

Analogous for "cells" image the following were obtained:

$$f_3(v) = 3.573 \cdot |v - 197| + 5.518 \cdot |v - 244.7| - 4.618 \cdot |v - 254|$$

$$f_4(v) = 3.156 \cdot |v - 197| + 13.502 \cdot |v - 243.3| - 8.972 \cdot |v - 246.2| - 3.212 \cdot |v - 254|$$

Their graphics are shown in Fig.1d and Fig.1f for "landsat", and also Fig.2d and Fig.2f for "cells". The enhanced images can be seen in Fig.1c, Fig.1e for the first image respectively in Fig.2c, Fig.2e for the second.

## 5. CONCLUSIONS

The paper presented a method for enhancing the gray level images. The method is based on point transforms defined by interpolation functions that are linear on some parts from the definition domain. These functions are determined by simple formulae, which need short calculus time. In establishing the interpolation points it was chosen an algorithm that considers the statistics properties of gray level images. Their points computing has a good convergence and requires few iterations. Future perspectives for the shown method could be:

1) the extension for color images;
2) fixing the interpolation points using the gray level fuzzification ([8]);
3) using the means of $k$ order for computing the points of interpolation ([5]);
4) the method approach in a logarithmical context ([4], [7]); this meaning a replacement of the classic operations from the real number algebra with operations from a logarithmical one [6].

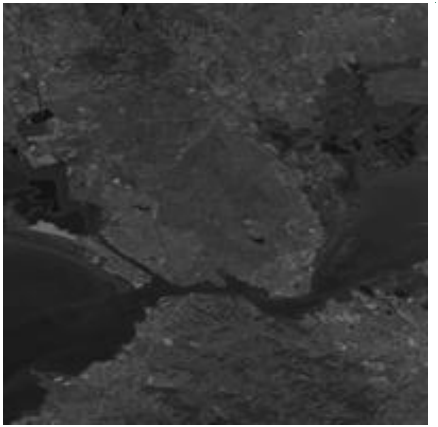

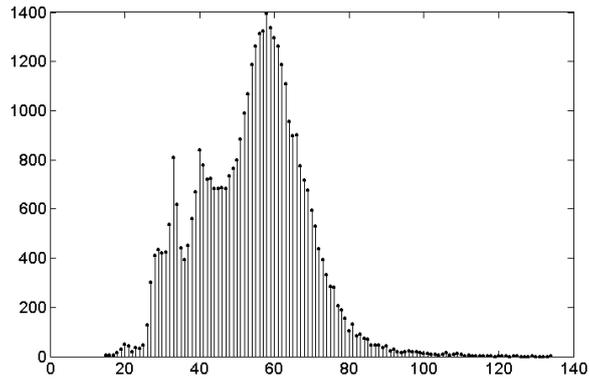

(a)                                               (b)

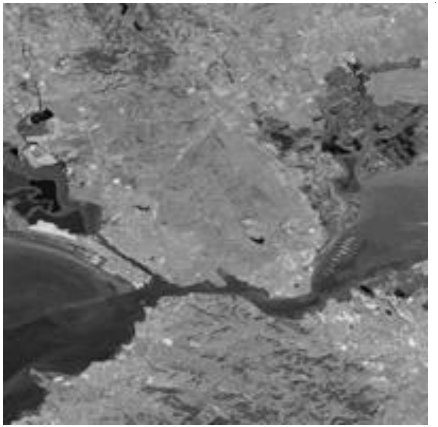

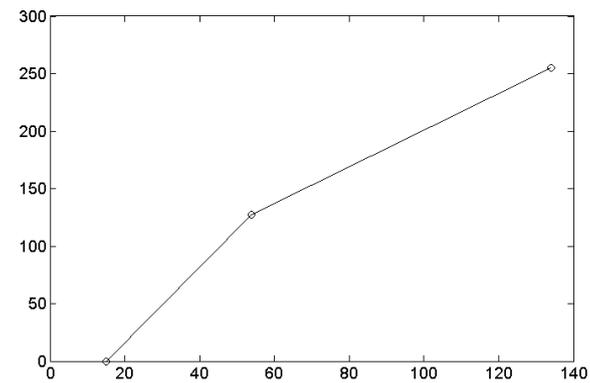

(c)                                               (d)

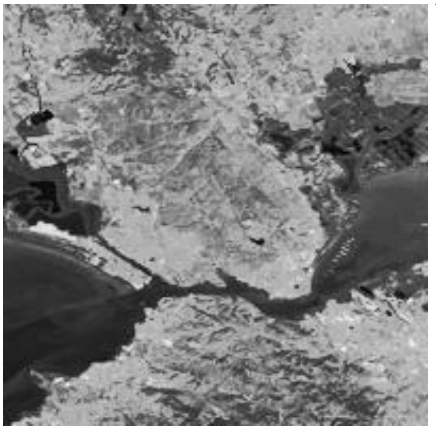

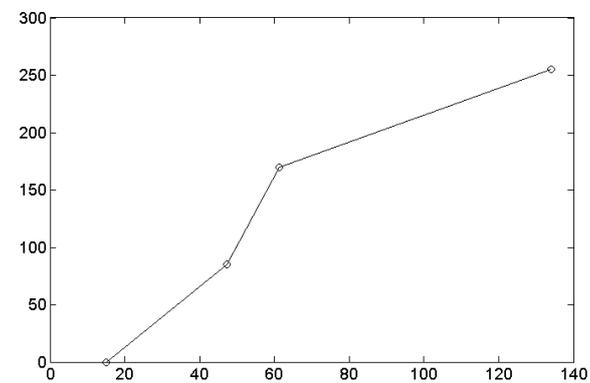

(e)                                               (f)

Fig. 1. (a) Original image "landsat" and (b) the gray level histogram,
(c) enhanced image and (d) the graphic of interpolation function for *n=3*,
(e) enhanced image and (f) the graphic of interpolation function for *n=4*.

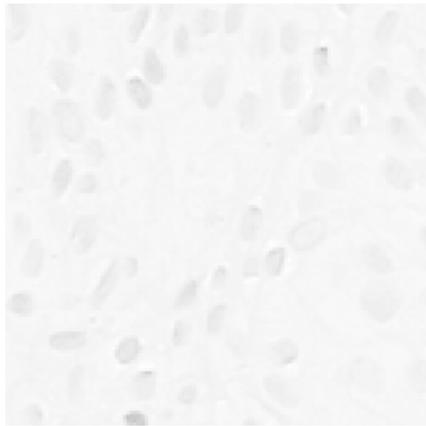

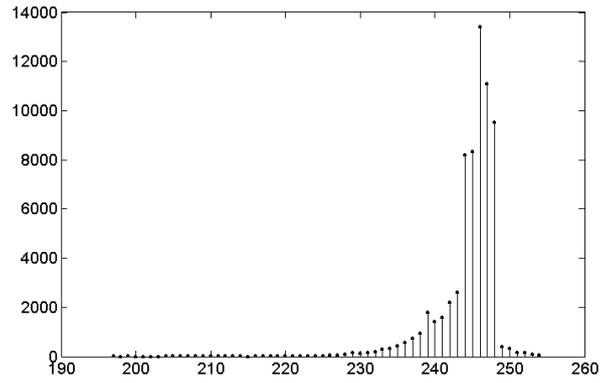

(a)

(b)

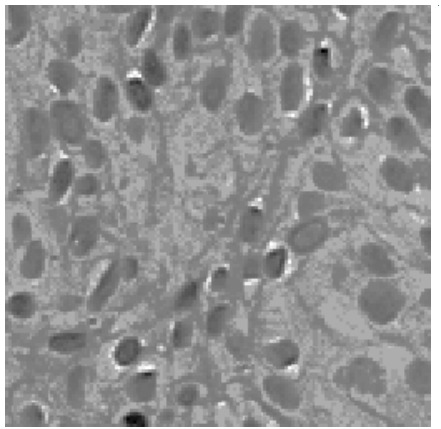

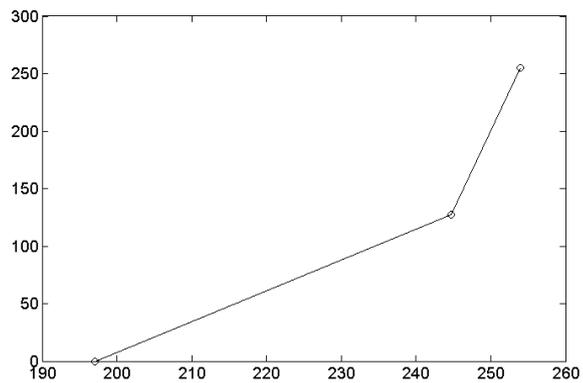

(c)

(d)

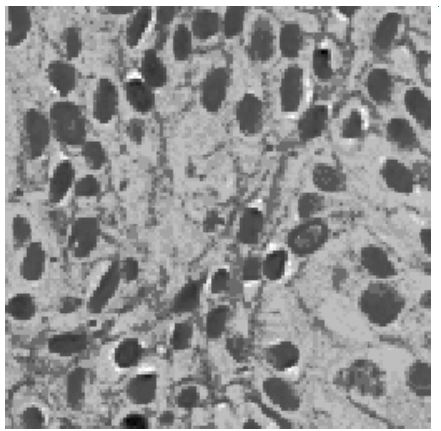

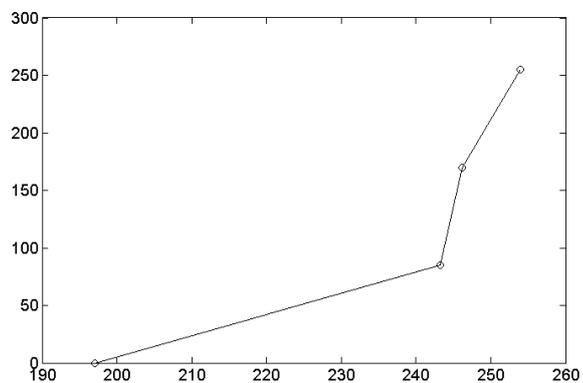

(e)

(f)

Fig. 2. (a) Original image "cells" and (b) the gray level histogram,
(c) enhanced image and (d) the graphic of interpolation function for *n=3*,
(e) enhanced image and (f) the graphic of interpolation function for *n=4*.